\documentclass{article}
\usepackage{spconf,amsmath,graphicx}
\usepackage{subfigure} 
\usepackage{algorithm}
\usepackage{algorithmic}
\usepackage{hyperref}
\usepackage{setspace}

\usepackage{amsthm}
\usepackage{amssymb}
\usepackage{bm}
\usepackage{multirow}
\usepackage{pifont}
\usepackage[capitalise]{cleveref}
\usepackage{color}
\usepackage[usenames,dvipsnames,svgnames,table]{xcolor}

\newcounter{mnote}

\newcommand{\eg}{e.g.\ }

\newcommand{\flabel}[1]{\label{fig:#1}}

\newcommand{\tlabel}[1]{\label{tab:#1}}
\newcommand{\elabel}[1]{\label{eq:#1}}

\newcommand{\fref}[1]{\cref{fig:#1}}
\newcommand{\sref}[1]{\cref{sec:#1}}
\newcommand{\tref}[1]{\cref{tab:#1}}

\newcommand*\idx[2][]
{ 
\def\next{#1}%
\ifx\empty\next
  (#2)
\else
  (#1, #2)
\fi
}
\newcommand*\elt[3][]
{ 
\def\next{#1}%
\ifx\empty\next
  #2\idx{#3}
\else
  #1\idx{#2,#3}
\fi
}
\newcommand*\pd[3][]
{ 
\def\next{#1}%
\ifx\empty\next
  \frac{\partial#2}{\partial #3}
\else
  \frac{{\partial^{#1} #2}}{\partial#3^{#1}}
\fi
}

\newcommand{\hfor}{\overrightarrow{h}}
\newcommand{\hback}{\overleftarrow{h}}
\newcommand{\igate}{i}
\newcommand{\fgate}{f}
\newcommand{\ogate}{o}
\newcommand{\state}{c}
\newcommand{\hiddenfn}{\mathcal{H}}

\newcommand{\wtmat}[2]{W_{#1 #2}}
\newcommand{\ihwts}{\wtmat{x}{h}}
\newcommand{\hhwts}{\wtmat{h}{h}}
\newcommand{\howts}{\wtmat{h}{y}}
\newcommand{\bias}[1]{b_{#1}}
\newcommand{\hbias}{\bias{h}}
\newcommand{\obias}{\bias{y}}
\newcommand{\seq}[1]{\boldsymbol #1}

\newcommand{\blank}{\varnothing}

\newcommand{\invble}{x}
\newcommand{\outvble}{y}
\newcommand{\targvble}{z}
\newcommand{\inseq}{\seq{\invble}}
\newcommand{\outseq}{\seq{\outvble}}

\newcommand{\targseq}{\seq{\targvble}}

\newcommand{\capt}[2]{\caption[#1]{#1#2}}

\newcommand{\figt}[5]
{
\begin{figure}[t]
\begin{center}
\includegraphics[width=#3\columnwidth]{figures/#1}
\end{center}
\capt{#4}{#5}
\flabel{#2}
\end{figure}
}

\title{Speech Recognition with Deep Recurrent Neural Networks}

\name{Alex Graves, Abdel-rahman Mohamed and Geoffrey Hinton}
\address{Department of Computer Science, University of Toronto}

\begin{document}
%
\maketitle

\begin{abstract}
Recurrent neural networks (RNNs) are a powerful model for sequential data.
End-to-end training methods such as Connectionist Temporal Classification make it possible to train RNNs for sequence labelling problems where the input-output alignment is unknown.
The combination of these methods with the Long Short-term Memory RNN architecture has proved particularly fruitful, delivering state-of-the-art results in cursive handwriting recognition.
However RNN performance in speech recognition has so far been disappointing, with better results returned by deep feedforward networks.
This paper investigates \emph{deep recurrent neural networks}, which combine the multiple levels of representation that have proved so effective in deep networks with the flexible use of long range context that empowers RNNs. 
When trained end-to-end with suitable regularisation, we find that deep Long Short-term Memory RNNs achieve a test set error of 17.7\% on the TIMIT phoneme recognition benchmark, which to our knowledge is the best recorded score.
\end{abstract}
\section{Introduction}
\label{sec:intro}
Neural networks have a long history in speech recognition, usually in combination with hidden Markov models~\cite{bourlard94hybrid,Zhu04tandem}.
They have gained attention in recent years with the dramatic improvements in acoustic modelling yielded by deep feedforward networks~\cite{5704567,6296526}.
Given that speech is an inherently dynamic process, it seems natural to consider recurrent neural networks (RNNs) as an alternative model.
HMM-RNN systems~\cite{robinson:1994} have also seen a recent revival~\cite{VinyalsICASSP12,maas2012denoisernn}, but do not currently perform as well as deep networks.

Instead of combining RNNs with HMMs, it is possible to train RNNs `end-to-end' for speech recognition~\cite{graves06icml,graves12supervised,graves12transducer}.
This approach exploits the larger state-space and richer dynamics of RNNs compared to HMMs, and avoids the problem of using potentially incorrect alignments as training targets.
The combination of Long Short-term Memory~\cite{hochreiter97lstm}, an RNN architecture with an improved memory, with end-to-end training has proved especially effective for cursive handwriting recognition~\cite{graves08nips,graves09nips}.
However it has so far made little impact on speech recognition.

RNNs are inherently deep in time, since their hidden state is a function of all previous hidden states.
The question that inspired this paper was whether RNNs could also benefit from depth in space; that is from stacking multiple recurrent hidden layers on top of each other, just as feedforward layers are stacked in conventional deep networks.
To answer this question we introduce \emph{deep Long Short-term Memory} RNNs and assess their potential for speech recognition.
We also present an enhancement to a recently introduced end-to-end learning method that jointly trains two separate RNNs as acoustic and linguistic models~\cite{graves12transducer}.
Sections~\ref{sec:rnn} and \ref{sec:training} describe the network architectures and training methods, \sref{experiments} provides experimental results and concluding remarks are given in~\sref{conclusion}.


\section{Recurrent Neural Networks}
\label{sec:rnn}

Given an input sequence $\inseq = (x_1,\ldots,x_T)$, a standard recurrent neural network (RNN) computes the hidden vector sequence $\seq{h} = (h_1,\ldots,h_T)$ and output vector sequence $\outseq = (y_1,\ldots,y_T)$ by iterating the following equations from $t=1$ to $T$:
\begin{align}
\elabel{rnn_hidden}
h_t &= \hiddenfn\left(\ihwts x_t + \hhwts h_{t-1} + \hbias \right)\\
y_t &= \howts h_t + \obias
\end{align} 
where the $W$ terms denote weight matrices (\eg $\ihwts$ is the input-hidden weight matrix), the $b$ terms denote bias vectors (\eg $\hbias$ is hidden bias vector) and $\hiddenfn$ is the hidden layer function.
 %



$\hiddenfn$ is usually an elementwise application of a sigmoid function.
However we have found that the Long Short-Term Memory (LSTM) architecture~\cite{hochreiter97lstm}, which uses purpose-built \emph{memory cells} to store information, is better at finding and exploiting long range context.
\fref{lstm} illustrates a single LSTM memory cell.
For the version of LSTM used in this paper~\cite{gers02peeps} $\hiddenfn$ is implemented by the following composite function:
\begin{align}
\igate_t &= \sigma\left(\wtmat{x}{\igate} x_t + \wtmat{h}{\igate} h_{t-1} + \wtmat{\state}{\igate} \state_{t-1}  + b_\igate\right)\\
\fgate_t &= \sigma\left(\wtmat{x}{\fgate} x_t + \wtmat{h}{\fgate} h_{t-1} + \wtmat{\state}{\fgate} \state_{t-1} + b_\fgate \right)\\
\state_t &= \fgate_t \state_{t-1} + \igate_t \tanh \left(\wtmat{x}{\state} x_t + \wtmat{h}{\state} h_{t-1} + b_\state \right)\\
\ogate_t &= \sigma\left(\wtmat{x}{\ogate} x_t + \wtmat{h}{\ogate} h_{t-1} + \wtmat{\state}{\ogate} \state_{t} + b_\ogate\right)\\
h_t &= \ogate_t \tanh(\state_t)
\end{align}
where $\sigma$ is the logistic sigmoid function, and $\igate$, $\fgate$, $\ogate$ and $\state$ are respectively the \emph{input gate}, \emph{forget gate}, \emph{output gate} and \emph{cell} activation vectors, all of which are the same size as the hidden vector $h$.
The weight matrices from the cell to gate vectors (\eg $\wtmat{s}{\igate}$) are diagonal, so element $m$ in each gate vector only receives input from element $m$ of the cell vector.

\figt{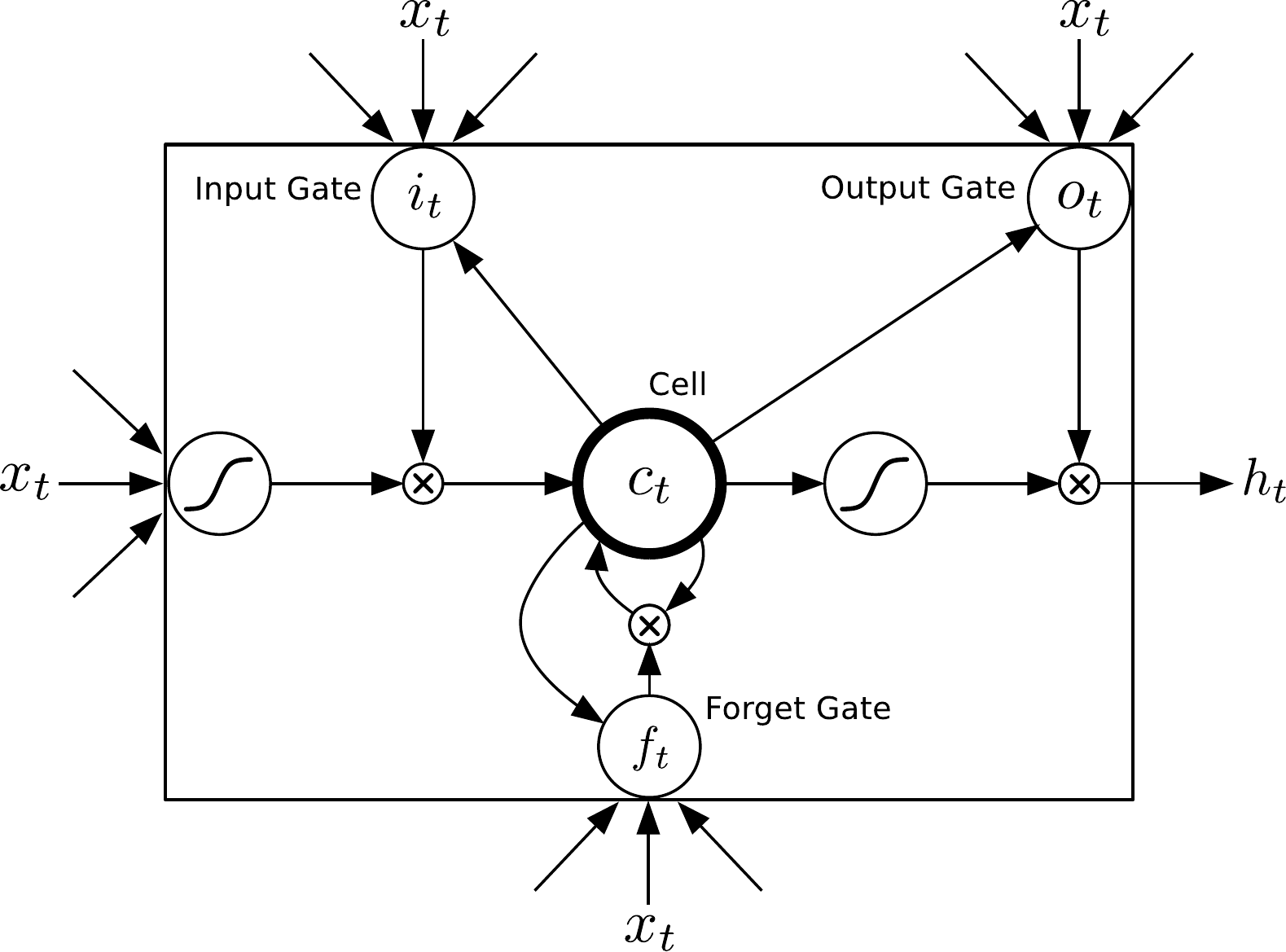}{lstm}{0.85}{Long Short-term Memory Cell}{}

One shortcoming of conventional RNNs is that they are only able to make use of previous context.
In speech recognition, where whole utterances are transcribed at once, there is no reason not to exploit future context as well.
Bidirectional RNNs (BRNNs)~\cite{schuster97bidirectional} do this by processing the data in both directions with two separate hidden layers, which are then fed forwards to the same output layer.
As illustrated in \fref{brnn}, a BRNN computes the \emph{forward} hidden sequence $\seq{\hfor}$, 
the \emph{backward} hidden sequence $\seq{\hback}$ 
and the output sequence $\outseq$ by iterating the backward layer from $t=T$ to $1$, the forward layer from $t=1$ to $T$ and then updating the output layer:
\figt{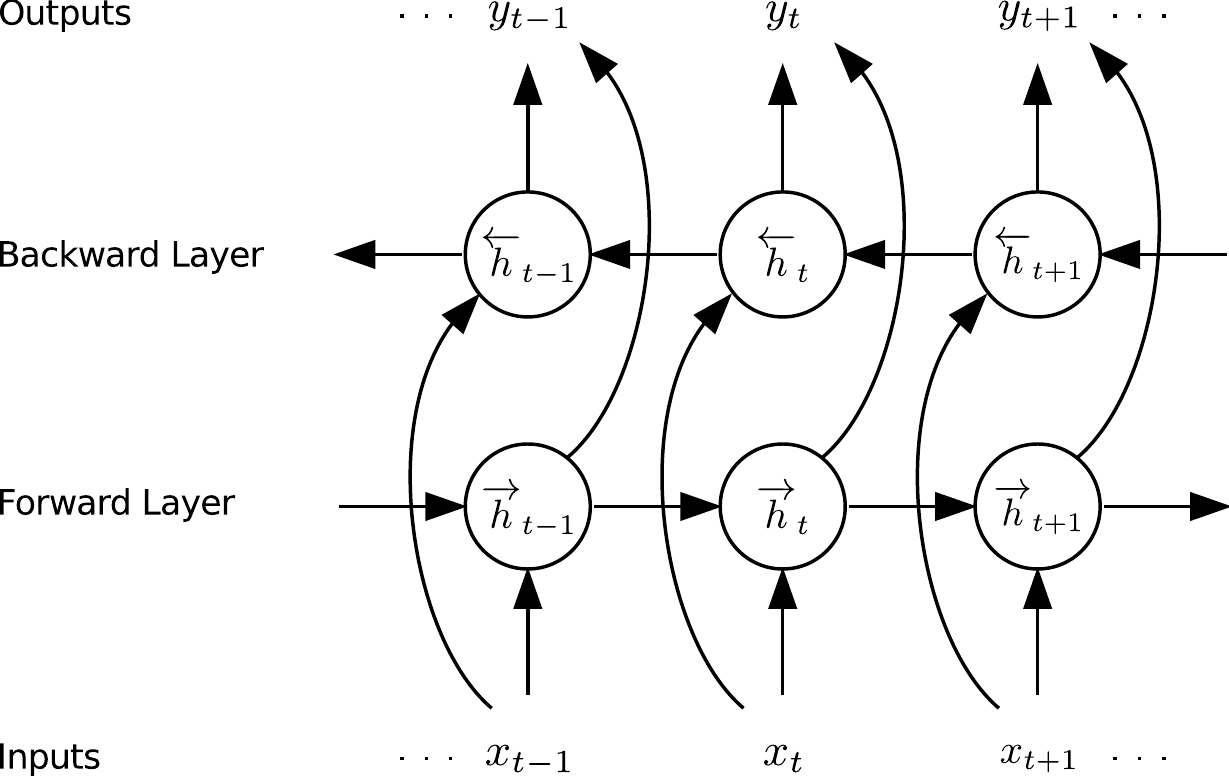}{brnn}{0.75}{Bidirectional RNN}{}
\begin{align}
\hfor_t &= \hiddenfn\left(\wtmat{x}{\hfor} x_t + \wtmat{\hfor}{\hfor} \hfor_{t-1} + \bias{\hfor} \right)\\
\hback_t &= \hiddenfn\left(\wtmat{x}{\hback} x_t + \wtmat{\hback}{\hback} \hback_{t+1} + \bias{\hback}\right)\\
y_t &= \wtmat{\hfor}{y} \hfor_t + \wtmat{\hback}{y} \hback_t + \obias
\end{align}
Combing BRNNs with LSTM gives bidirectional LSTM~\cite{graves05nn}, which can access long-range context in both input directions.

A crucial element of the recent success of hybrid HMM-neural network systems is the use of \emph{deep} architectures, which are able to build up progressively higher level representations of acoustic data.
\emph{Deep RNNs} can be created by stacking multiple RNN hidden layers on top of each other, with the output sequence of one layer forming the input sequence for the next. Assuming the same hidden layer function is used for all $N$ layers in the stack, the hidden vector sequences $\seq{h}^n$ are iteratively computed from $n=1$ to $N$ and $t=1$ to $T$:
\begin{align}
\elabel{deep_rnn_hidden}
h^n_t &= \hiddenfn\left(W_{h^{n-1}h^{n}} h^{n-1}_t + W_{h^{n}h^{n}} h^n_{t-1} + \hbias^n \right)
\end{align}
where we define $\seq{h}^0 = \inseq$. The network outputs $y_t$ are 
\begin{align}
y_t &= \wtmat{h^N}{y} h^N_t + \obias
\end{align}
Deep bidirectional RNNs can be implemented by replacing each hidden sequence $\seq{h}^n$ with the forward and backward sequences $\seq{\hfor}^n$ and $\seq{\hback}^n$, and ensuring that every hidden layer receives input from both the forward and backward layers at the level below.
If LSTM is used for the hidden layers we get deep bidirectional LSTM, the main architecture used in this paper.
As far as we are aware this is the first time deep LSTM has been applied to speech recognition, and we find that it yields a dramatic improvement over single-layer LSTM.

\section{Network Training}
\label{sec:training}
We focus on end-to-end training, where RNNs learn to map directly from acoustic to phonetic sequences.
One advantage of this approach is that it removes the need for a predefined (and error-prone) alignment to create the training targets.
The first step is to to use the network outputs to parameterise a differentiable distribution $\Pr(\outseq|\inseq)$ over all possible phonetic output sequences $\outseq$ given an acoustic input sequence $\inseq$.
The log-probability $\log\Pr(\targseq|\inseq)$ of the target output sequence $\targseq$ can then be differentiated with respect to the network weights using backpropagation through time~\cite{rumelhart88backprop}, and the whole system can be optimised with gradient descent.
We now describe two ways to define the output distribution and hence train the network.
We refer throughout to the length of $\inseq$ as $T$, the length of $\targseq$ as $U$, and the number of possible phonemes as $K$.

\subsection{Connectionist Temporal Classification}
The first method, known as Connectionist Temporal Classification (CTC)~\cite{graves06icml,graves12supervised}, uses a softmax layer to define a separate output distribution $\Pr(k|t)$ at every step $t$ along the input sequence.
This distribution covers the $K$ phonemes plus an extra blank symbol $\blank$ which represents a non-output (the softmax layer is therefore size $K+1$).
Intuitively the network decides whether to emit any label, or no label, at every timestep.
Taken together these decisions define a distribution over alignments between the input and target sequences.
CTC then uses a forward-backward algorithm to sum over all possible alignments and determine the normalised probability $\Pr(\targseq|\inseq)$ of the target sequence given the input sequence~\cite{graves06icml}.
Similar procedures have been used elsewhere in speech and handwriting recognition to integrate out over possible segmentations~\cite{zweig09scarf,senior95nips}; however CTC differs in that it ignores segmentation altogether and sums over single-timestep label decisions instead.

RNNs trained with CTC are generally bidirectional, to ensure that every $\Pr(k|t)$ depends on the entire input sequence, and not just the inputs up to $t$.
In this work we focus on deep bidirectional networks, with $\Pr(k|t)$ defined as follows:
\begin{align}
y_{t} &= W_{\hfor^N y} \hfor^N_{t} + W_{\hback^N y} \hback^N_{t} + b_y\\
\Pr(k|t) &= \frac{\exp(y_{t}[k])}{\sum_{k'=1}^K{\exp(y_{t}[k'])}},
\end{align}
where $y_{t}[k]$ is the $k^{th}$ element of the length $K+1$ unnormalised output vector $y_{t}$, and $N$ is the number of bidirectional levels.

\subsection{RNN Transducer}
CTC defines a distribution over phoneme sequences that depends only on the acoustic input sequence $\inseq$.
It is therefore an acoustic-only model.
A recent augmentation, known as an \emph{RNN transducer}~\cite{graves12transducer} combines a CTC-like network with a separate RNN that predicts each phoneme given the previous ones, thereby yielding a jointly trained acoustic and language model.
Joint LM-acoustic training has proved beneficial in the past for speech recognition~\cite{Mohamed10investigationof,5495227}.

Whereas CTC determines an output distribution at every input timestep, an RNN transducer determines a separate distribution $\Pr(k|t,u)$ for every \emph{combination} of input timestep $t$ and output timestep $u$.
As with CTC, each distribution covers the $K$ phonemes plus $\blank$.
Intuitively the network `decides' what to output depending both on where it is in the input sequence and the outputs it has already emitted.
For a length $U$ target sequence $\targseq$, the complete set of $TU$ decisions jointly determines a distribution over all possible alignments between $\inseq$ and $\targseq$, which can then be integrated out with a forward-backward algorithm to determine $\log\Pr(\targseq|\inseq)$~\cite{graves12transducer}.

In the original formulation $\Pr(k|t,u)$ was defined by taking an `acoustic' distribution $\Pr(k|t)$ from the CTC network, a `linguistic' distribution $\Pr(k|u)$ from the prediction network, then multiplying the two together and renormalising.
An improvement introduced in this paper is to instead feed the hidden activations of both networks into a separate feedforward \emph{output network}, whose outputs are then normalised with a softmax function to yield $\Pr(k|t,u)$.
This allows a richer set of possibilities for combining linguistic and acoustic information, and appears to lead to better generalisation.
In particular we have found that the number of deletion errors encountered during decoding is reduced.

Denote by $\seq{\hfor^N}$ and $\seq{\hback^N}$ the uppermost forward and backward hidden sequences of the CTC network, and by $\seq{p}$ the hidden sequence of the prediction network.
At each $t,u$ the output network is implemented by feeding $\seq{\hfor^N}$ and $\seq{\hback^N}$ to a linear layer to generate the vector $l_t$, then feeding $l_t$ and $p_u$ to a $\tanh$ hidden layer to yield $h_{t,u}$, and finally feeding $h_{t,u}$ to a size $K+1$ softmax layer to determine $\Pr(k|t,u)$:
\begin{align}
l_{t} &= W_{\hfor^N l} \hfor^N_t + W_{\hback^N l} \hback^N_t + b_{l}\\
h_{t, u} &= \tanh\left(W_{l h} l_{t, u} + W_{p b} p_u + b_{h}\right)\\
y_{t,u} &= W_{h y} h_{t, u} + b_y\\
\Pr(k|t,u) &= \frac{\exp(y_{t,u}[k])}{\sum_{k'=1}^K{\exp(y_{t,u}[k'])}},
\end{align}
where $y_{t,u}[k]$ is the $k^{th}$ element of the length $K+1$ unnormalised output vector.
For simplicity we constrained all non-output layers to be the same size ($|\hfor^n_t| = |\hback^n_t| = |p_u| = |l_t| = |h_{t, u}|)$; however they could be varied independently.

RNN transducers can be trained from random initial weights.
However they appear to work better when initialised with the weights of a pretrained CTC network and a pretrained next-step prediction network (so that only the output network starts from random weights).
The output layers (and all associated weights) used by the networks during pretraining are removed during retraining.
In this work we pretrain the prediction network 
on the phonetic transcriptions of the audio training data; however for large-scale applications it would make more sense to pretrain on a separate text corpus.

\subsection{Decoding}
RNN transducers can be decoded with beam search~\cite{graves12transducer} to yield an n-best list of candidate transcriptions.
In the past CTC networks have been decoded using either a form of best-first decoding known as \emph{prefix search}, or by simply taking the most active output at every timestep~\cite{graves06icml}.
In this work however we exploit the same beam search as the transducer, with the modification that the output label probabilities $\Pr(k|t,u) $ do not depend on the previous outputs (so $\Pr(k|t,u) = \Pr(k|t)$).
We find beam search both faster and more effective than prefix search for CTC.
Note the n-best list from the transducer was originally sorted by the \emph{length normalised} log-probabilty $\log \Pr(\outseq)/|\outseq|$; in the current work we dispense with the normalisation (which only helps when there are many more deletions than insertions) and sort by $\Pr(\outseq)$.

\subsection{Regularisation}
Regularisation is vital for good performance with RNNs, as their flexibility makes them prone to overfitting.
Two regularisers were used in this paper: early stopping and \emph{weight noise} (the addition of Gaussian noise to the network weights during training~\cite{chuen96noise}).
Weight noise was added once per training sequence, rather than at every timestep. 
Weight noise tends to `simplify' neural networks, in the sense of reducing the amount of information required to transmit the parameters~\cite{hinton93bitsback,graves11nips}, which improves generalisation.

\section{Experiments}
\label{sec:experiments}

Phoneme recognition experiments were performed on the TIMIT corpus~\cite{timit}.
The standard 462 speaker set with all SA records removed was used for training, and a separate development set of 50 speakers was used for early stopping. 
Results are reported for the 24-speaker core test set.
The audio data was encoded using a Fourier-transform-based filter-bank with 40 coefficients (plus energy) distributed on a mel-scale, together with their first and second temporal derivatives.
Each input vector was therefore size 123.
The data were normalised so that every element of the input vectors had zero mean and unit variance over the training set. 
All 61 phoneme labels were used during training and decoding (so $K=61$), then mapped to 39 classes for scoring~\cite{lee89timit39}.
Note that all experiments were run only once, so the variance due to random weight initialisation and weight noise is unknown.

As shown in \tref{timit}, nine RNNs were evaluated, varying along three main dimensions: the training method used (CTC, Transducer or pretrained Transducer), the number of hidden levels (1--5), and the number of LSTM cells in each hidden layer. 
Bidirectional LSTM was used for all networks except CTC-3l-500h-tanh, which had $\tanh$ units instead of LSTM cells, and CTC-3l-421h-uni where the LSTM layers were unidirectional.
All networks were trained using stochastic gradient descent, with learning rate $10^{-4}$, momentum $0.9$ and random initial weights drawn uniformly from $[-0.1,0.1]$.
All networks except CTC-3l-500h-tanh and PreTrans-3l-250h were first trained with no noise and then, starting from the point of highest log-probability on the development set, retrained with Gaussian weight noise ($\sigma=0.075$) until the point of lowest phoneme error rate on the development set.
PreTrans-3l-250h was initialised with the weights of CTC-3l-250h, along with the weights of a phoneme prediction network (which also had a hidden layer of 250 LSTM cells), both of which were trained without noise, retrained with noise, and stopped at the point of highest log-probability.
PreTrans-3l-250h was trained from this point with noise added.
CTC-3l-500h-tanh was entirely trained without weight noise because it failed to learn with noise added.
Beam search decoding was used for all networks, with a beam width of 100.

The advantage of deep networks is immediately obvious, with the error rate for CTC dropping from 23.9\% to 18.4\% as the number of hidden levels increases from one to five.
The four networks CTC-3l-500h-tanh, CTC-1l-622h, CTC-3l-421h-uni and CTC-3l-250h all had approximately the same number of weights, but give radically different results.
The three main conclusions we can draw from this are (a) LSTM works much better than $\tanh$ for this task, (b) bidirectional LSTM has a slight advantage over unidirectional LSTMand (c) depth is more important than layer size (which supports previous findings for deep networks~\cite{5704567}).
%
%
Although the advantage of the transducer is slight when the weights are randomly initialised, it becomes more substantial when pretraining is used.

\begin{table}
\centering
\capt{TIMIT Phoneme Recognition Results.}{ `Epochs' is the number of passes through the training set before convergence. `PER' is the phoneme error rate on the core test set.}
\tlabel{timit}
\vskip 0.15in
\begin{center}
\begin{small}
\begin{sc}\begin{tabular}{llll}
\hline
Network & Weights & Epochs & PER\\
\hline
CTC-3l-500h-tanh & 3.7M & 107 & 37.6\%\\
CTC-1l-250h & 0.8M & 82 & 23.9\%\\
CTC-1l-622h & 3.8M & 87 & 23.0\%\\
CTC-2l-250h & 2.3M & 55 & 21.0\%\\
CTC-3l-421h-uni & 3.8M & 115 & 19.6\%\\
CTC-3l-250h & 3.8M & 124 & 18.6\%\\
CTC-5l-250h & 6.8M & 150 & 18.4\%\\
Trans-3l-250h & 4.3M & 112 & 18.3\%\\
\textbf{PreTrans-3l-250h} & \textbf{4.3M} & \textbf{144} & \textbf{17.7\%}\\
\hline
\end{tabular}
\end{sc}
\end{small}
\end{center}
\vskip -0.1in
\end{table}

\figt{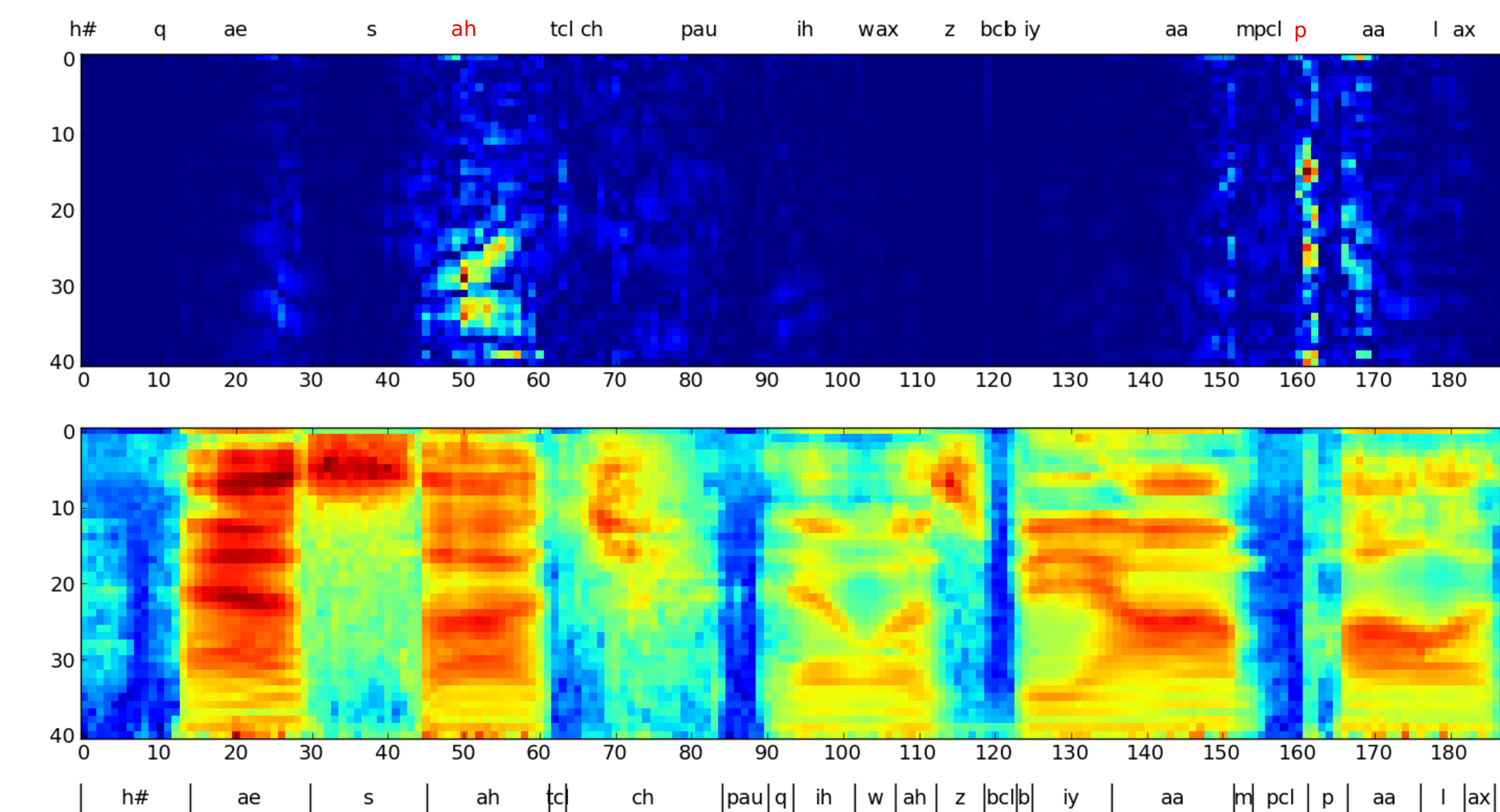}{sensitivity}{1}{Input Sensitivity of a deep CTC RNN.}{ The heatmap (top) shows the derivatives of the `ah' and `p' outputs printed in red with respect to the filterbank inputs (bottom). The TIMIT ground truth segmentation is shown below. 
Note that the sensitivity extends to surrounding segments; this may be because CTC (which lacks an explicit language model) attempts to learn linguistic dependencies from the acoustic data.}



\section{Conclusions and future work}
\label{sec:conclusion}

We have shown that the combination of deep, bidirectional Long Short-term Memory RNNs with end-to-end training and weight noise gives state-of-the-art results in phoneme recognition on the TIMIT database.
An obvious next step is to extend the system to large vocabulary speech recognition.
Another interesting direction would be to combine frequency-domain convolutional neural networks~\cite{mohamed12cnn} with deep LSTM.

\clearpage
\begin{spacing}{0.97}
\bibliographystyle{IEEEbib}
\bibliography{refs}
\end{spacing}
\end{document}